\documentclass[runningheads]{llncs}

\usepackage{accv}

\usepackage{accvabbrv}

\usepackage{graphicx}
\usepackage{booktabs}
\usepackage{multirow}

\usepackage[accsupp]{axessibility}  

\usepackage[pagebackref,breaklinks,colorlinks,citecolor=accvblue]{hyperref}
\usepackage{hyperref}

\usepackage{orcidlink}
\usepackage{xcolor}
\definecolor{blue}{RGB}{0,0,0}

\begin{document}

\title{Beyond Imperfections: A Conditional Inpainting Approach  for End-to-End Artifact Removal in VTON and Pose Transfer} 

\titlerunning{Artifact Removal}

\author{Aref Tabatabaei\inst{1} \and
Zahra Dehghanian\inst{1} \and
Maryam Amirmazlaghani\inst{1}}
\institute{Amirkabir University of Technology }

\maketitle

\begin{abstract}

Artifacts often degrade the visual quality in virtual try-on (VTON) and pose transfer applications, \textcolor{blue}{negatively} impacting user experience. This study introduces a novel conditional inpainting technique to detect and \textcolor{blue}{eliminate these} distortions, \textcolor{blue}{thereby enhancing } image aesthetics. Our work is the first to \textcolor{blue}{propose} an end-to-end framework \textcolor{blue}{specifically targeting this issue, supported by a newly developed dataset of artifacts in VTON and pose transfer tasks, which includes masks that mark affected regions.}. Experimental results \textcolor{blue}{demonstrate} that our method not only effectively removes artifacts but also significantly \textcolor{blue}{improves} the visual quality of the final images, \textcolor{blue}{establishing} a new benchmark in computer vision and image processing. 
\end{abstract}

\label{sec:intro}
\section{Introduction}
\begin{figure}[t]
  \centering
  \includegraphics[width=0.6\columnwidth]{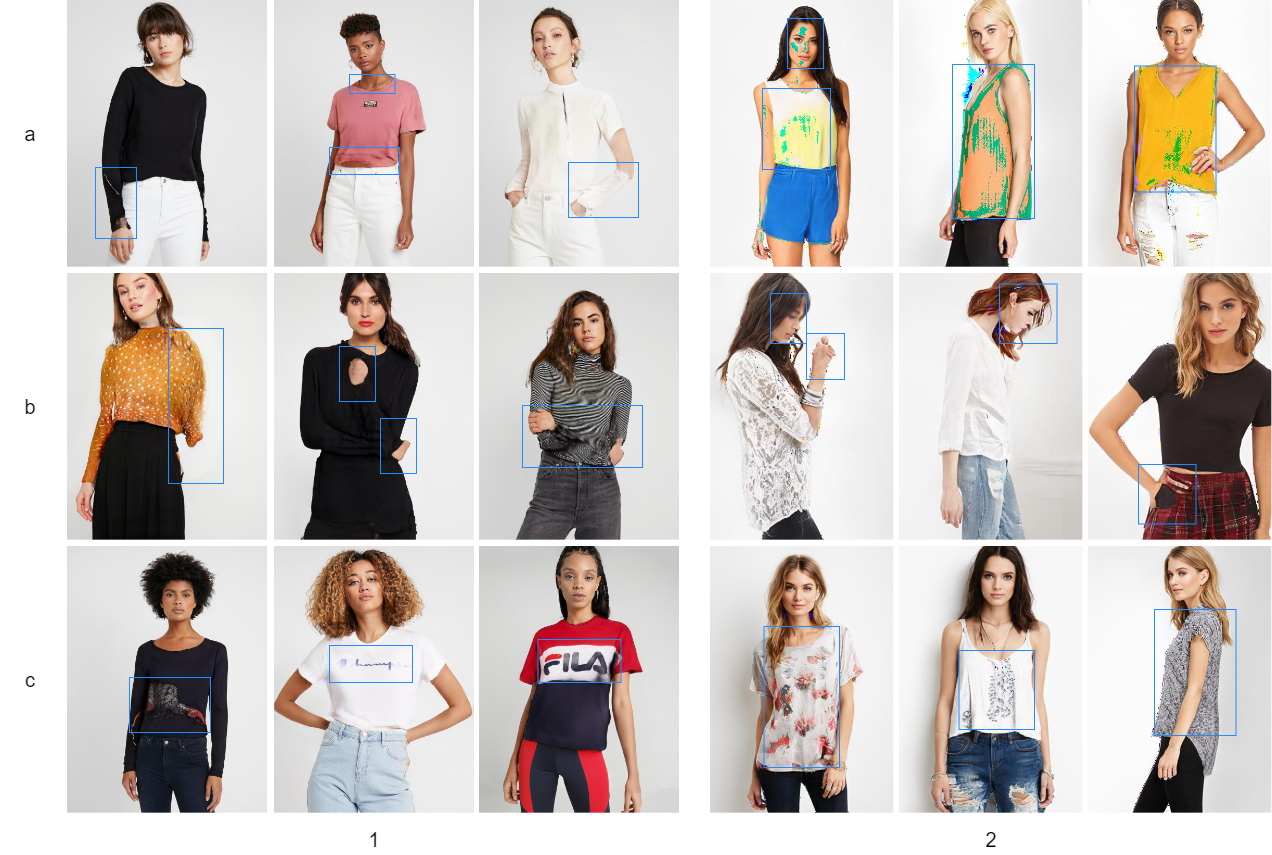}

   \caption{\textbf{Sample of Artifacts in Our datasets}: 
Rows a), b), and c) showcase distinct artifact types: color and texture, deformation in body parts, and cloth design. Images in our datasets are further classified by dataset origin: 1) VDI dataset (VTON-HD Distorted Images) and 2) DDI dataset (Deepfashion Distorted Images). Details on datasets will be discussed on Section \ref{sec:datasets}.}
   \label{fig1}
\end{figure}

Virtual try-on (VTON) and pose transfer tasks have \textcolor{blue}{seen} substantial advancements in recent years; however, persistent artifacts \textcolor{blue}{such as} unnatural distortions in the generated \textcolor{blue}{images} remain a critical challenge (See figure \ref{fig1}). 

This paper proposes a novel method for artifact removal using Stable Diffusion \cite{rombach2022high} inpainting, \textcolor{blue}{to improve} image realism and visual fidelity. Our approach incorporates ControlNet \cite{zhang2023adding} and IP-Adapter \cite{ye2023ip-adapter} for fine-tuning the inpainting process. ControlNet allows conditioning image inputs to guide image generation, while IP-Adapter integrates image features into a text-to-image diffusion model. 

Additionally, we introduce an automatic artifact detection mechanism to efficiently identify and rectify distortions. We also present two specialized datasets \textcolor{blue}{—DDI for VTON and VDI for pose transfer tasks—containing} distorted images and corresponding masks from real-world scenarios. \textcolor{blue}{Experimental results validate } the effectiveness of our method in removing artifacts and enhancing image quality compared to existing approaches. All code and datasets will be publicly available.

We summarize our contributions as follows: 

\begin{itemize}
    \item Artifact Detection: A comprehensive detection mechanism leveraging multi-strategy approaches for a wide range of artifact types.
    \item Innovative Artifact Removal Technique: Our method dynamically adjusts the inpainting process based on the artifact nature, ensuring optimal correction.  
    \item Specialized New Datasets: Custom datasets for VTON and pose transfer artifact removal, facilitating targeted development and assessment. 
\end{itemize}

 \section{Related Work}
In the field of virtual try-on and pose transfer, significant progress has been made using image-based techniques to overlay clothing onto target figures while maintaining pose integrity. Early work, such as VITON \cite{han2018viton}, utilized encoder-decoder frameworks and TPS transformations to improve clothing warping. Subsequent enhancements incorporated learnable TPS modules and body information \cite{wang2018toward, fele2022c}, often relying on GANs for high-quality image generation \cite{goodfellow2014generative}. Challenges in misalignment were addressed through methods like normalization, distillation \cite{ge2021parser}, and semantic-aware discriminators \cite{morelli2022dress}, leading to more accurate and realistic outcomes.

GANs are powerful generative networks that have been widely applied across various domains \cite{kang2023scaling,dehghanian2023spot}, and have revolutionized tasks such as human pose transfer by facilitating the generation of realistic images starting with Ma et al.'s model \cite{ma2017pose} that combined source images with target pose heatmaps. These techniques evolved from multi-stage processes emphasizing geometric models \cite{ren2020deep} to single-stage methods focusing on efficiency \cite{zhang2021pise}.

Diffusion models have emerged as powerful tools for various image synthesis tasks, including virtual try-on. Foundational work by Ho et al. \cite{ho2020denoising} and Rombach et al. \cite{rombach2022high} laid the groundwork for high-quality image generation within latent spaces. Recent advancements like TryOnDiffusion \cite{zhu2023tryondiffusion} and LADI-VTON \cite{morelli2023ladi} have further refined these techniques, though challenges such as dataset collection and detail loss persist \cite{gou2023taming}. However, limitations in handling clothing texture and shape changes remain, with recent work exploring UV maps and flow-based methods \cite{sarkar2021style, ren2021combining} for more nuanced control. Our approach aims to further enhance pose transfer quality by integrating advanced inpainting techniques to address common artifacts.

\section{Methodology}
\label{sec:blind}
\begin{figure}[t]
  \centering
  \includegraphics[width=0.8\columnwidth]{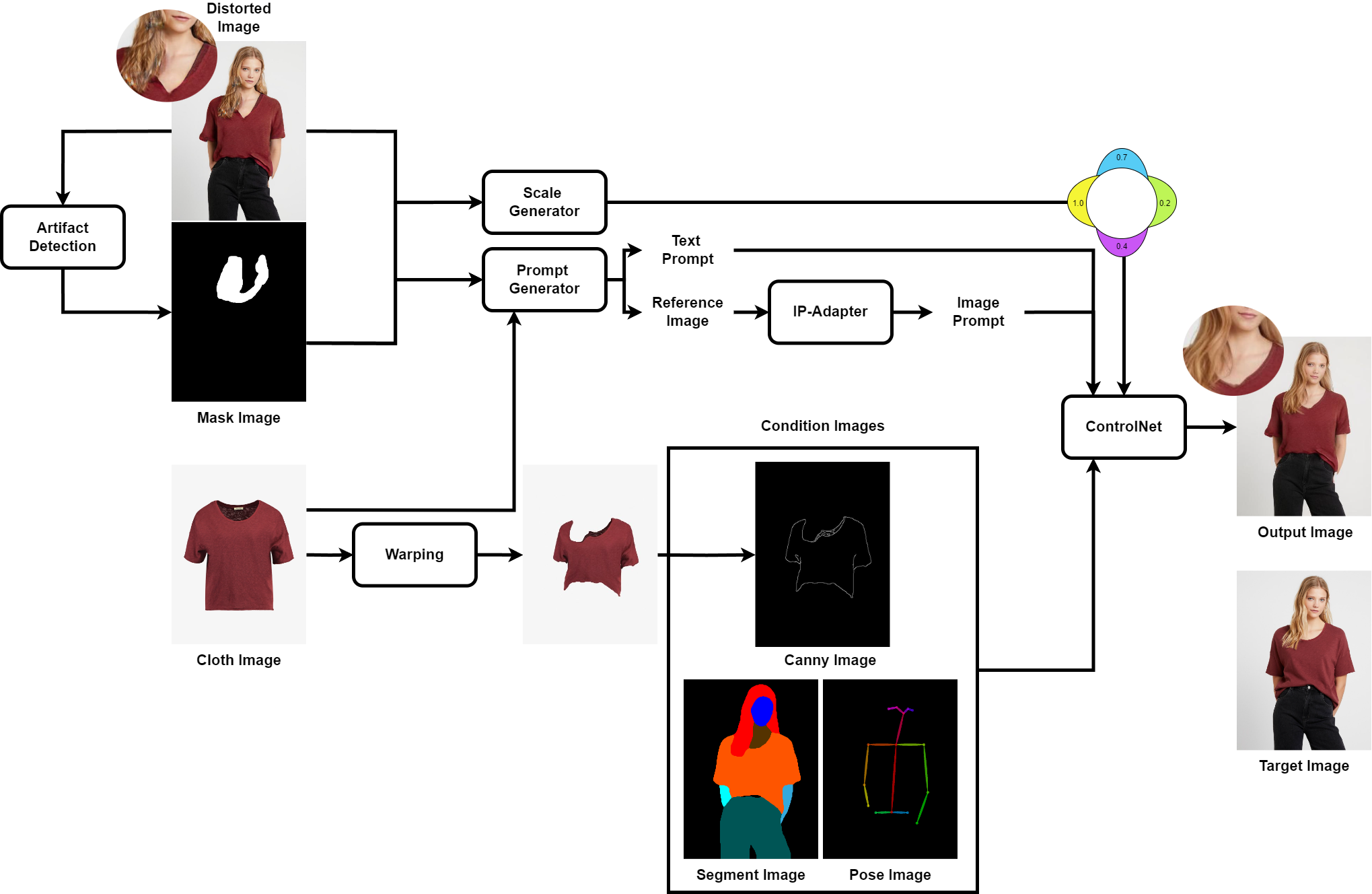}
   \caption{{End-to-End Artifact Removal Model Architecture Overview: In the shown example, our artifact detection model identifies that the hair and collar of the shirt are distorted and generates a mask for the affected region. Following this, conditions and prompts are generated, and simultaneously, our scale generator model calculates impact scales for each condition. In this example, the Canny condition (blue) and IP-Adapter (yellow) exhibit higher impact, while the pose (green) and segmentation (purple) have lower impact. }}
   \label{fig2}
\end{figure}

In figure \ref{fig2}, the structure of our model is visually presented, offering an insightful overview of the system's components and their interactions. In pose transfer and virtual try-on (VTON) tasks, output images often contain artifacts, resulting in distorted images. To address this, we propose an automatic artifact detection method that generates a mask image highlighting the artifact regions in the input image. The target cloth image in VTON or the human image in both tasks can be utilized as a condition image. For the conditioning images used in the ControlNet \cite{zhang2023adding} model, we extract the Canny edge image \cite{canny1986computational} of the cloth, the semantic map, and the pose image. We then generate a text prompt for the inpainting module, complemented by an image prompt derived from the input image using the IP-Adapter model \cite{ye2023ip-adapter}. Finally, the Stable Diffusion model is employed to regenerate the artifact regions within the distorted image. We'll explore the workflow of each part step by step. 





\subsection{Automatic Artifact Detection}
\label{sec: artifact detection}
In virtual try-on and pose transfer tasks, artifacts can significantly detract from the visual quality of the \textcolor{blue}{generated} images (See figure \ref{fig1}). Existing methods for artifact localization are \textcolor{blue}{limited, prompting us to develop} a novel approach to detect artifacts using input images or human features. We classify artifacts into three main categories: \textbf{Color and Texture Artifacts}, involving uneven color distribution or texture issues; \textbf{Deformation Artifacts}, which include abnormalities in body parts leading to unnatural poses; and \textbf{Distortion in Cloth Design} causing deviations from the intended design or structure. This classification helps in understanding the challenges in artifact detection and removal, providing a structured approach to enhancing image fidelity in these applications.

Our automatic artifact detection methodology employs a comprehensive four-pronged strategy, each aspect meticulously designed to uncover specific types of artifacts. Firstly, the YOLO model \cite{redmon2016you} is utilized to precisely identify human features such as hands and faces within the image. If the model's confidence level falls below a predetermined threshold for these features, the corresponding area is identified as an artifact. Subsequently, a mask is generated to precisely delineate these regions for subsequent corrective measures.

Secondly, through Color Palette Comparison \cite{gonzalez2009digital}, we distill the image's color scheme to a simpler palette. This step is crucial for spotlighting stark color deviations, which often hint at underlying color-related distortions, thus offering a straightforward method to pinpoint these issues.

The third tactic, Canny Edge Matching \cite{canny1986computational}, is applied specifically to clothing items within the image. By warping the target cloth to match the target pose and then performing Canny edge detection on both the warped items and the one in input image, discrepancies in the edge mappings serve as clear indicators of potential distortions, thereby flagging these areas as problematic.

Lastly, Pose Keypoints Matching addresses the alignment of poses in virtual try-on (VTON) and pose transfer tasks. This involves a direct comparison between the intended pose and the pose as rendered in the generated image. Disparities here, indicative of pose misalignment, are immediately classified as artifacts, necessitating additional focus and refinement in the subsequent inpainting process.

\begin{figure}[t]
  \centering
  \includegraphics[width=0.5\columnwidth]{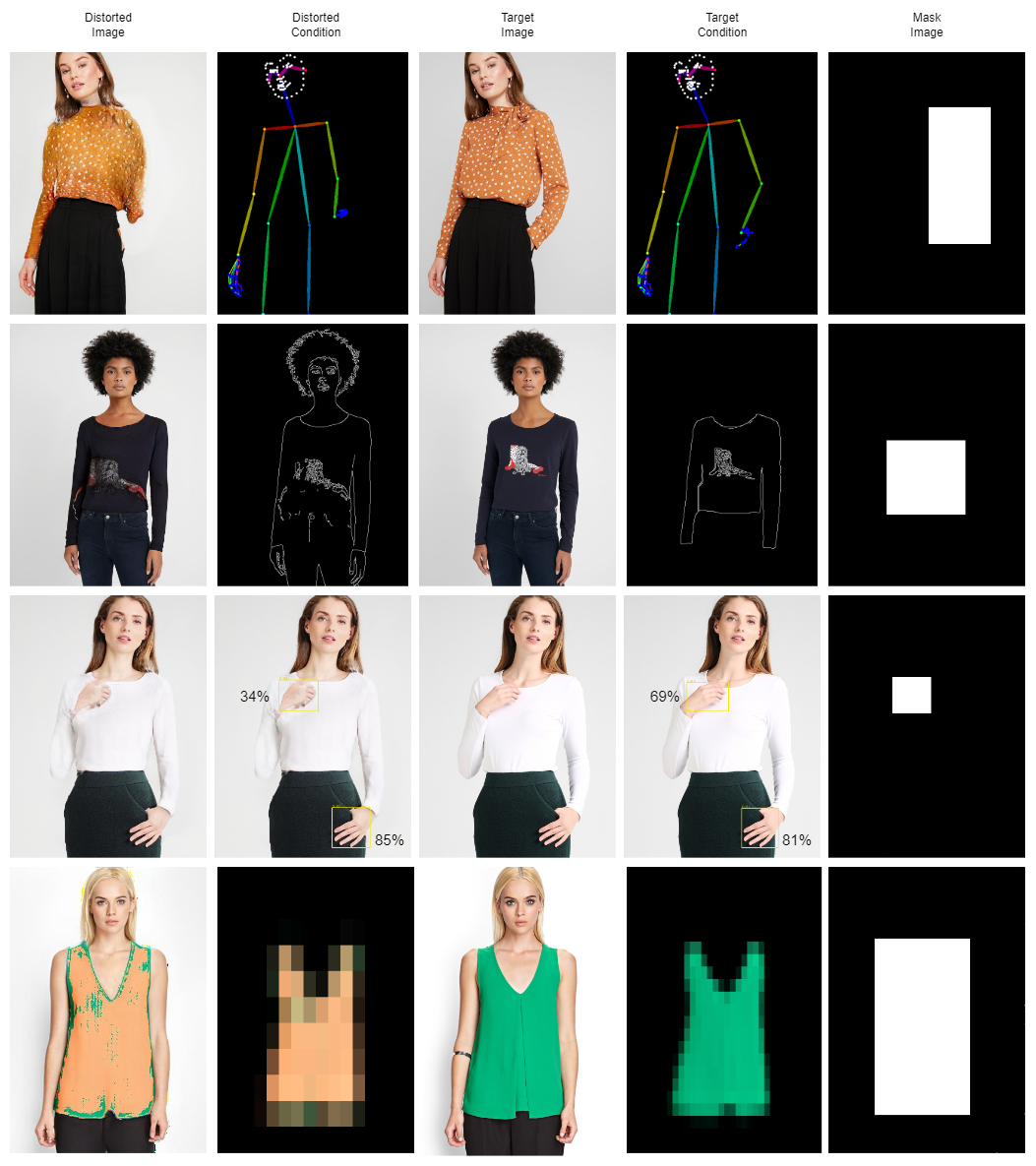}

   \caption{  {Automatic Artifact Detection: Comparison of distorted and target conditions, including pose image, Canny edge detection image, color palette and YOLO detection}}
   \label{fig3}
\end{figure}

Figure \ref{fig3} illustrates the effectiveness of this multi-faceted approach, showing how each technique identifies and marks potential artifacts 
\subsection{Artifact Restoration Inpainter}

In our architecture, the ControlNet \cite{zhang2023adding} plays a \textcolor{blue}{crucial role in ensuring accuracy} in the inpainting process. It accepts diverse conditioning images to guide Stable Diffusion in generating realistic and artifact-free \textcolor{blue}{outputs}. \textcolor{blue}{To optimize ControlNet’s guidance, we incorporate multiple conditioning inputs with fine-tuned weights to direct it effectively.}

These \textcolor{blue}{conditioning inputs} are generated at the same time. The prompt generator pulls out text details related to the visual content and feeds them into the model. Since distortions vary greatly, the model needs \textcolor{blue}{distinct conditioning images} to accurately address and fix all types of distortions. The scale generator works with the distorted image and its mask to \textcolor{blue}{determine the appropriate influence of each conditioning input} in the inpainting process of the ControlNet.

In simpler terms, after finding the messed-up parts of the image, we tell our system where to focus its fixing efforts. Our inpainting section needs three things to start: a text description of what's in the picture, pictures that show what everything should look like, and a guide on how important each of these pictures is for the fix. The text description helps the system understand the picture better. Because not all mess-ups are the same, the system looks at different reference pictures to know how to fix them. Then, it decides how much to rely on each reference picture for the best fix. This way, we make sure the system fixes the image accurately, making it look good again. In the next subsections we will delve into each module.

\subsubsection{Condition Images}
As said before, we used several image conditions which are generated as below:
\begin{itemize}
\item\textbf{Canny Edge Condition.}
    A warping process \cite{zhang2020human} is employed to align the extracted cloth image with the target pose, and at last step, the Canny edge detection technique is applied to the cloth image, extracting relevant structural features.

To visually demonstrate the impact of warping and the Canny Edge condition, figure \ref{fig4} highlights how the combination of warping and the Canny Image condition contributes to the effective alignment and feature extraction crucial for artifact removal in virtual try-on and pose transfer scenarios.
\begin{figure}[t]
  \centering
  \includegraphics[width=0.6\columnwidth]{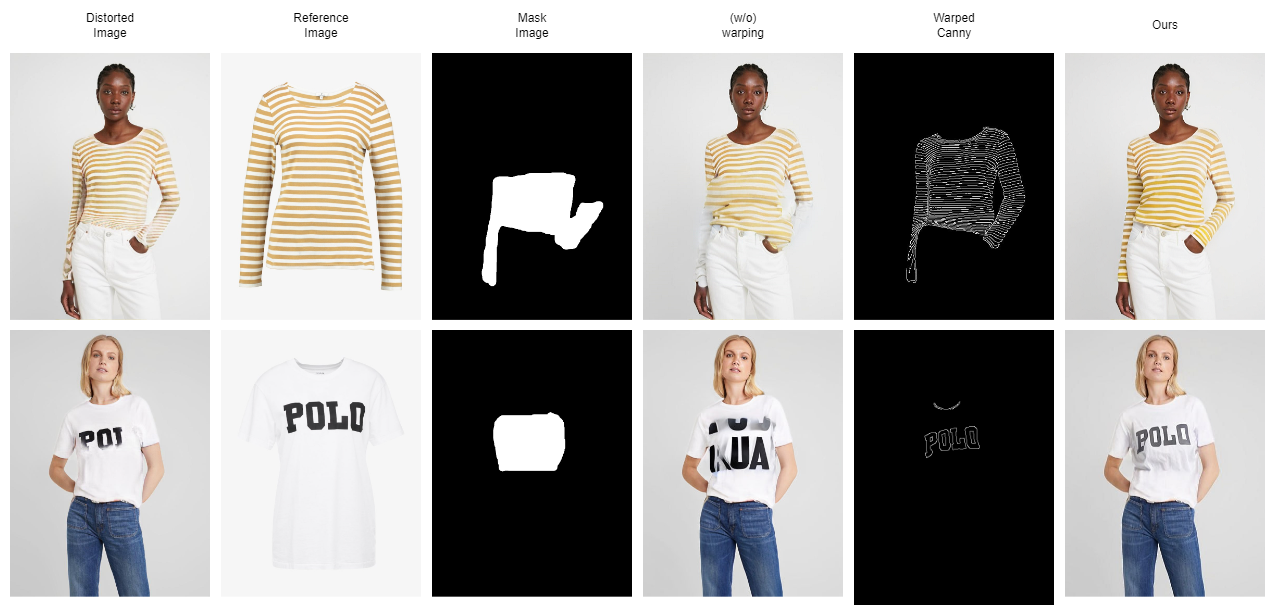}

   \caption{  {Usage of the Canny Condition: Highlighting the importance of warping to align the Canny edges within the mask region in the distorted image}}
   \label{fig4}
\end{figure}
\item\textbf{Pose Keypoints Condition.}
The Pose Keypoints condition is employed differently in pose transfer and virtual try-on (VTON) tasks:
In pose transfer, pose keypoints are extracted directly from the input pose image but in VTON, pose keypoints are obtained using the OpenPose model \cite{cao2017realtime} applied to the source image.
To visually illustrate the impact of the Pose Keypoints condition, figure \ref{fig5} showcases the transformation of an image with the integration of pose keypoints. Notably, in ControlNet \cite{zhang2023adding}, the initial training of the human pose condition involved simple poses without keypoints for hands. Recognizing the importance of capturing finger details, we performed a fine-tuning of the pose condition with more complex poses that incorporate hands keypoints. Specifically, this condition significantly aids in reconstructing images where artifacts stem from deformation types, especially in hands or legs, demonstrating its critical role in enhancing image quality and accuracy in these complex areas. The visual results in figure \ref{fig5} highlight the discernible impact of the fine-tuning process on the effectiveness of the Pose Keypoints condition. 
 \begin{figure}[t]
  \centering
  \includegraphics[width=0.6\columnwidth]{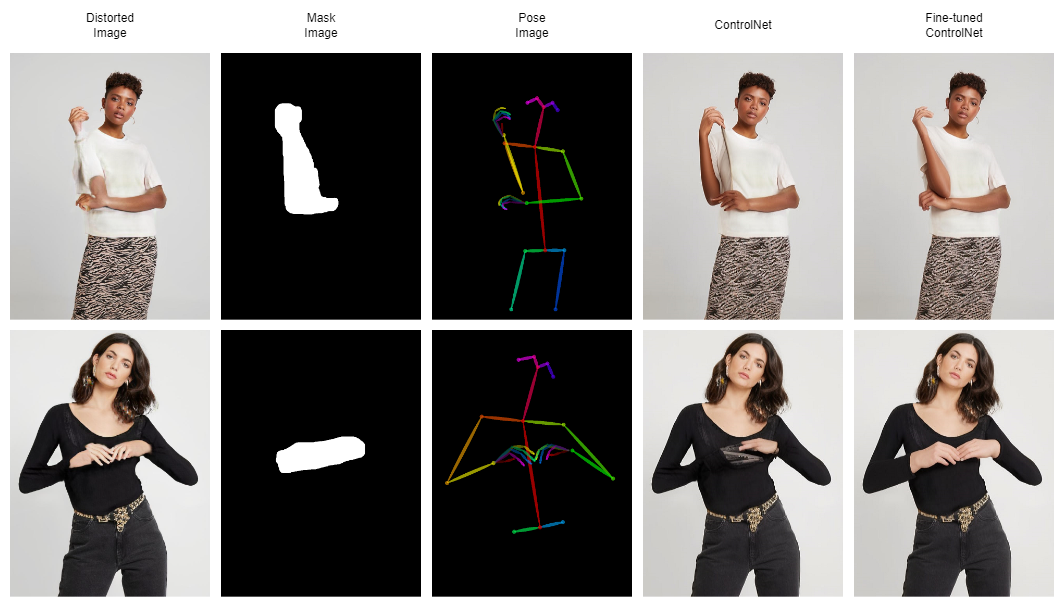}

   \caption{  {Usage of the Pose Condition: Comparison between the base ControlNet \cite{zhang2023adding} and the finetuned version}}
   \label{fig5}
\end{figure}
\item\textbf{Semantic Segmentation Condition}.
Despite the absence of a target segmentation map during test time in virtual try-on (VTON) and pose transfer tasks, various models \cite{lee2022high, li2023virtual, zhang2020human} exist to predict it based on given inputs. Leveraging this capability, we employed the segmentation map condition to address artifacts arising from confused or distorted boundaries of human body parts. During the training of ControlNet \cite{zhang2023adding} for semantic segmentation maps, a dataset containing diverse elements was initially utilized. However, recognizing the specific focus required for our task, we fine-tuned the model using human parsing data from the VTON-HD dataset  \cite{choi2021viton}.
In figure \ref{fig6}, the impact of the segmentation map condition is evident, showcasing its effectiveness in addressing artifacts related to human body part boundaries. Additionally, the visual results illustrate the discernible impact of our fine-tuning process on the performance of this condition.
\begin{figure}[t]
  \centering
  \includegraphics[width=0.6\columnwidth]{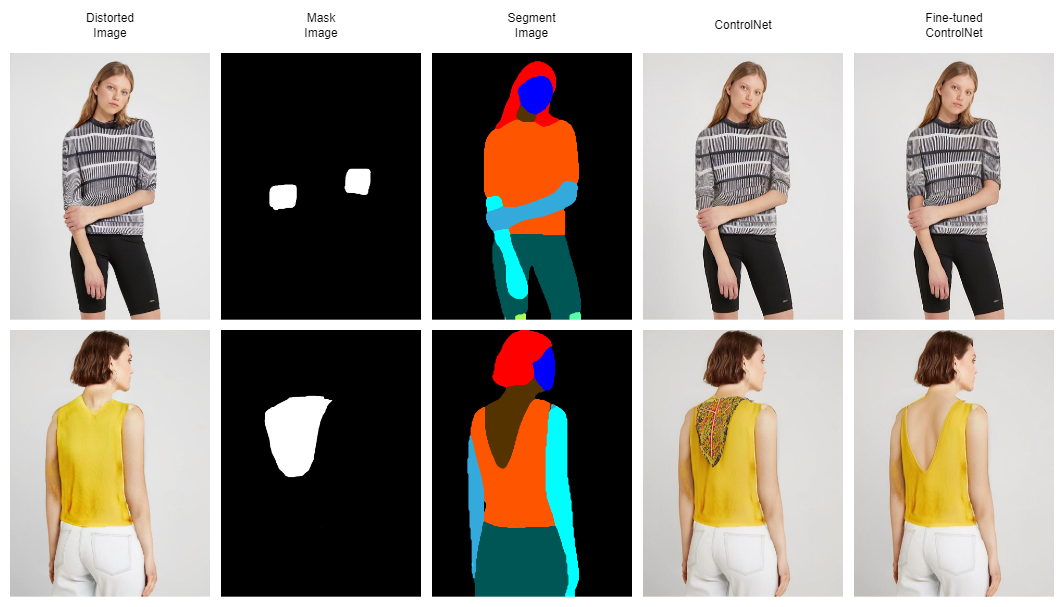}

   \caption{  {Usage of the Segmentation Condition: Comparison between the base ControlNet \cite{zhang2023adding} and the finetuned version}}
   \label{fig6}
\end{figure}
\end{itemize}

\subsubsection{IP-Adapter}
\begin{figure}[t]
  \centering
  \includegraphics[width=0.5\columnwidth]{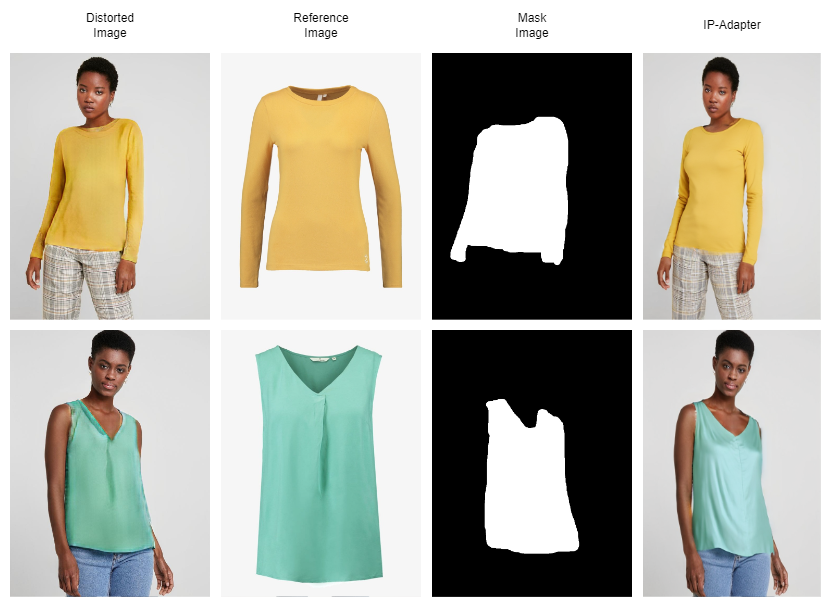}

   \caption{  {Usage of IP-Adapter: Addressing color artifacts to improve overall image quality and fidelity}}
   \label{fig7}
\end{figure}

As outlined in the Related Work section, the Image Prompt Adapter (IP-Adapter) \cite{ye2023ip-adapter} is a pivotal component designed to empower a pretrained text-to-image diffusion model. Comprising an image encoder and adapted modules with decoupled cross-attention, IP-Adapter enables the generation of images guided by specific image prompts. Particularly instrumental in reconstructing color artifacts, figure \ref{fig7} vividly illustrates the effectiveness of IP-Adapter in the removal of such artifacts.
\subsubsection{Prompt Generator}

\begin{figure}[t]
  \centering
  \includegraphics[width=0.6\columnwidth]{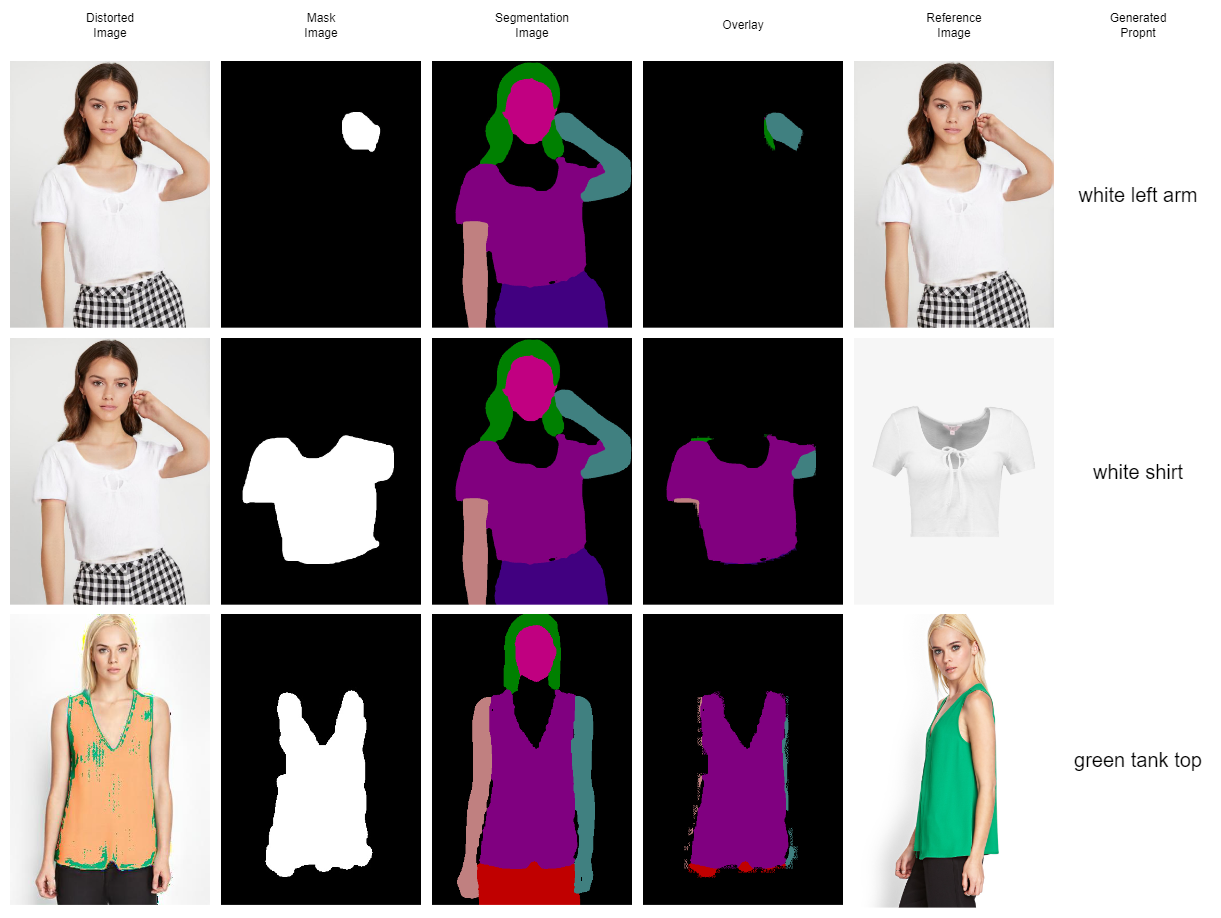}

   \caption{  {Automatic Prompt Generation Process: This process relies on the mask region in the distorted image and its underlying content. A text prompt is generated, and a reference (condition) image is selected for the image prompt.}}
   \label{fig8}
\end{figure}

As demonstrated in \cite{tabatabaei2024no} To effectively guide the inpainting process, we need to use a detailed text prompt along with an appropriate masking technique to achieve better results.
Here we pass input image along with mask image into our prompt generator model. 
We use Self Correction for Human Parsing model \cite{li2019selfcorrection} to accurately localizing the regions of the body in the input image that require inpainting. Following mask region identification, the BLIP \cite{li2022blip} model is used in extracting textual information relevant to the inpainting task. if the identification phase tells us that the mask section is cloth, we will pass the cloth image and if it's anywhere else, we will use the distorted image as a reference image. Then we will pass the reference image with the fixed prompt into the BLIP model to generate textual details that complement the visual content for us. This text-based information enriches the inpainting process by offering valuable insights such as color descriptions, context, or any other pertinent details. This text prompt, combined with the image prompt captured by the IP-Adapter, is then passed on to the inpainting process.  \\
figure \ref{fig8} highlights the identified mask regions along with the textual information generated by BLIP. This combined prompt serves as a comprehensive guide for Stable Diffusion, fostering a nuanced and accurate inpainting process. 

\subsubsection{Scale Generator}
A dynamic and adaptive scaling mechanism that enables ControlNet to respond differentially based on the specific artifact types, contributing to a nuanced and effective artifact removal process. It is trained as a context-aware ResNet model \cite{he2016deep} after training other modules. This specialized model, fundamental to our artifact removal strategy, is trained using a carefully curated dataset that will be elaborated in the next section. It takes two inputs—the distorted image and the corresponding mask—and outputs numerical values corresponding to the impact scales of different conditions in ControlNet.

\begin{figure}[t]
  \centering
  \includegraphics[width=0.6\columnwidth]{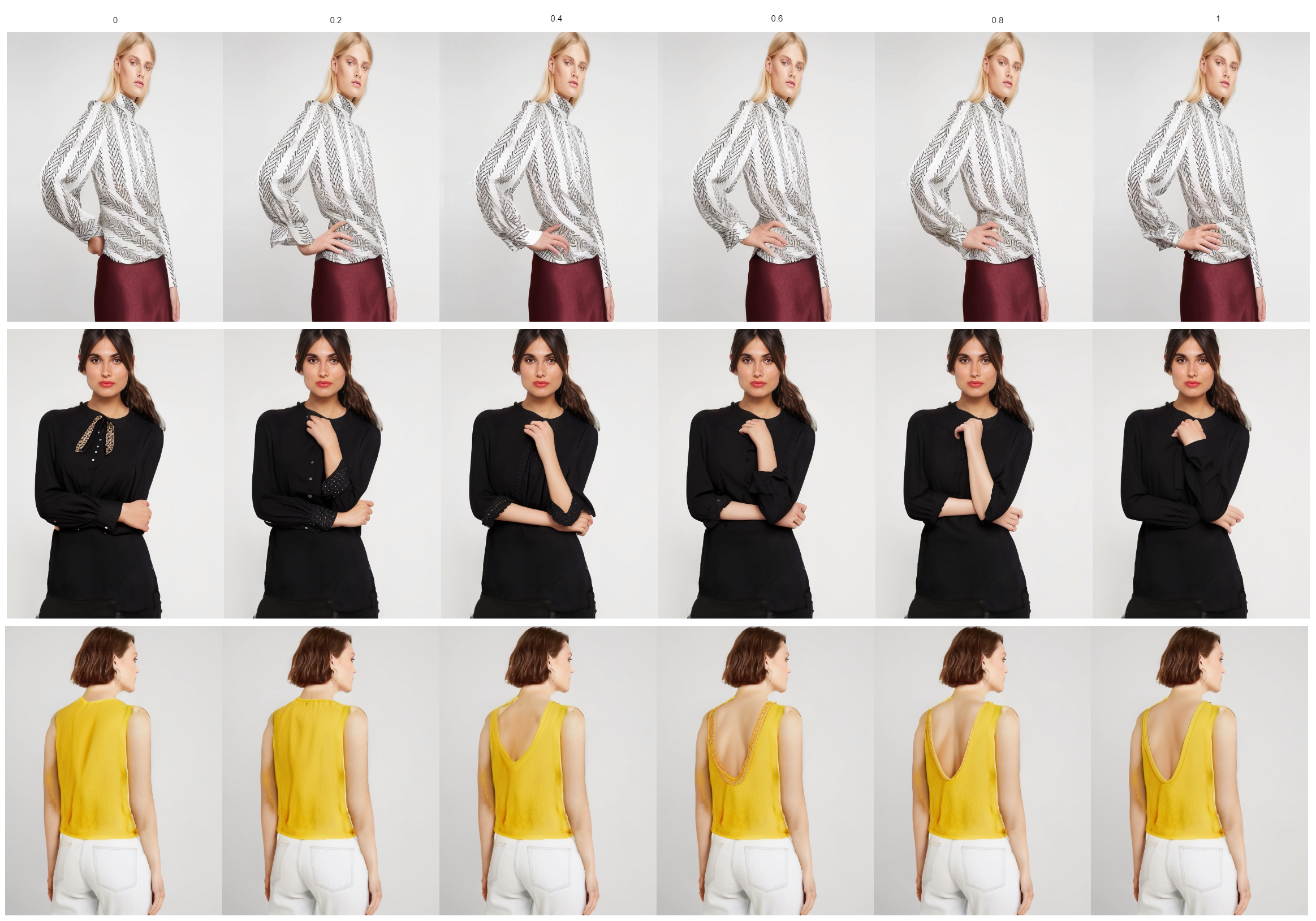}

   \caption{ Showcasing the Importance of Choosing the Proper Condition Scale: Demonstrating that neither a higher nor a lower scale is always optimal; the appropriate scale must be selected based on the specific context and desired outcome}
   \label{fig9}
\end{figure}

As demonstrated in figure \ref{fig9} scales do not always achieve optimal results at their highest or lowest ranges. Depending on the nature of the artifact, varying the scale can significantly enhance the effectiveness of artifact removal. 

\section{Datasets}
\label{sec:datasets}
In this paper, we introduce two novel datasets specifically tailored for artifact localization and removal in virtual try-on and pose transfer tasks (See figure \ref{fig10a}). 

\subsection{DDI}
The first, Deepfashion Distorted Images (DDI), utilizes the WOMEN-Blouses-Shirts in Deepfashion dataset \cite{liuLQWTcvpr16DeepFashion} to create 3673 instances of distorted images for pose transfer, each accompanied by a target image, a corresponding mask image, and multiple reference images from same person in different poses.
The resolution of distorted images and the masks is 352*512 and the resolution of the target image and reference images is 750*1101.

To generate distorted images, we input images into the NTED model \cite{ren2022neural}- a pose transfer model- along with the same pose to reconstruct the images. Subsequently, we manually pick the distorted one and generate mask images, capturing all mentioned variants of artifacts.

\subsection{VDI}
The second dataset, VTON-HD Distorted Images (VDI), is tailored for virtual try-on (VTON) tasks. The utilization of the VTON-HD dataset \cite{choi2021viton} primarily involves images focusing on upper clothing, with human subjects typically positioned from the full face down to above the knees. To generate the distorted images the HR-VTON model \cite{lee2022high}, a high-resolution VTON model is used. We input images of individuals wearing the same clothing for reconstruction and then select the distorted one and generate mask images. VDI comprises 2032 instances, each maintaining the resolution of VTON-HD dataset images 768*1024.

\begin{figure}[!ht]
    \centering
    \begin{subfigure}[t]{0.54\textwidth}
        \centering
        \includegraphics[width=\linewidth]{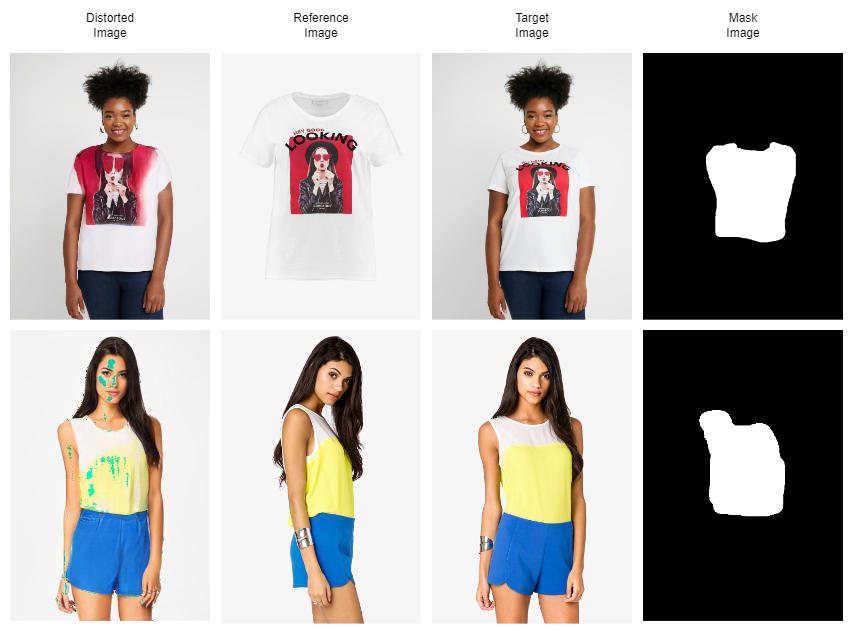}
        \caption{Datasets: The first row is from the VDI dataset, with the reference image as a cloth image. The second row is from the DDI dataset, with the reference image as the same person in a different pose.}
        \label{fig10a}
    \end{subfigure}
    \hspace{0.5em} 
    \begin{subfigure}[t]{0.39\textwidth}
        \centering
        \includegraphics[width=\linewidth]{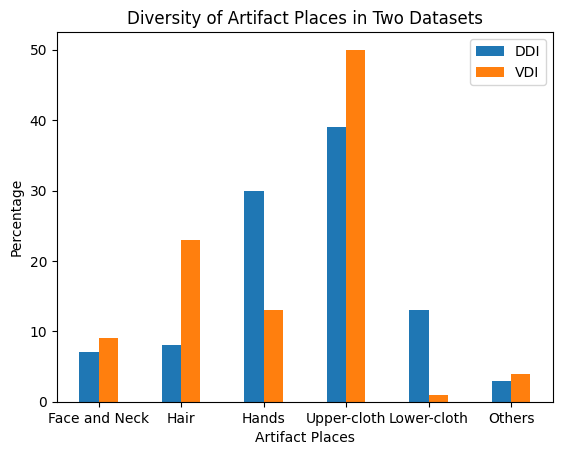}
        \caption{Diversity of artifact placements in two datasets.}
        \label{fig10b}
    \end{subfigure}
    \caption{Illustration of datasets and artifact diversity.}
    \label{fig:combined}
    \vspace{-1em} 
\end{figure}

\subsection{Diversity of Artifact Places}
By understanding the prevalent artifacts in virtual try-on (VTON) and pose transfer models, we curated datasets reflecting the diversity of these common issues. The histograms presented showcase the distribution of artifacts across different regions such as Face and Neck, Hair, Hands, Upper-cloth, Lower-cloth, and Others in both datasets (Figure \ref{fig10b}). Sometimes, there are multiple artifacts present in a single image. In such cases, we separate them for each distorted image with multiple mask images, ensuring that each artifact is individually annotated.








\section{EXPERIMENTAL EVALUATION}
\label{sec: evaluation}
\begin{figure}[t]
  \centering
  \includegraphics[width=0.6\columnwidth]{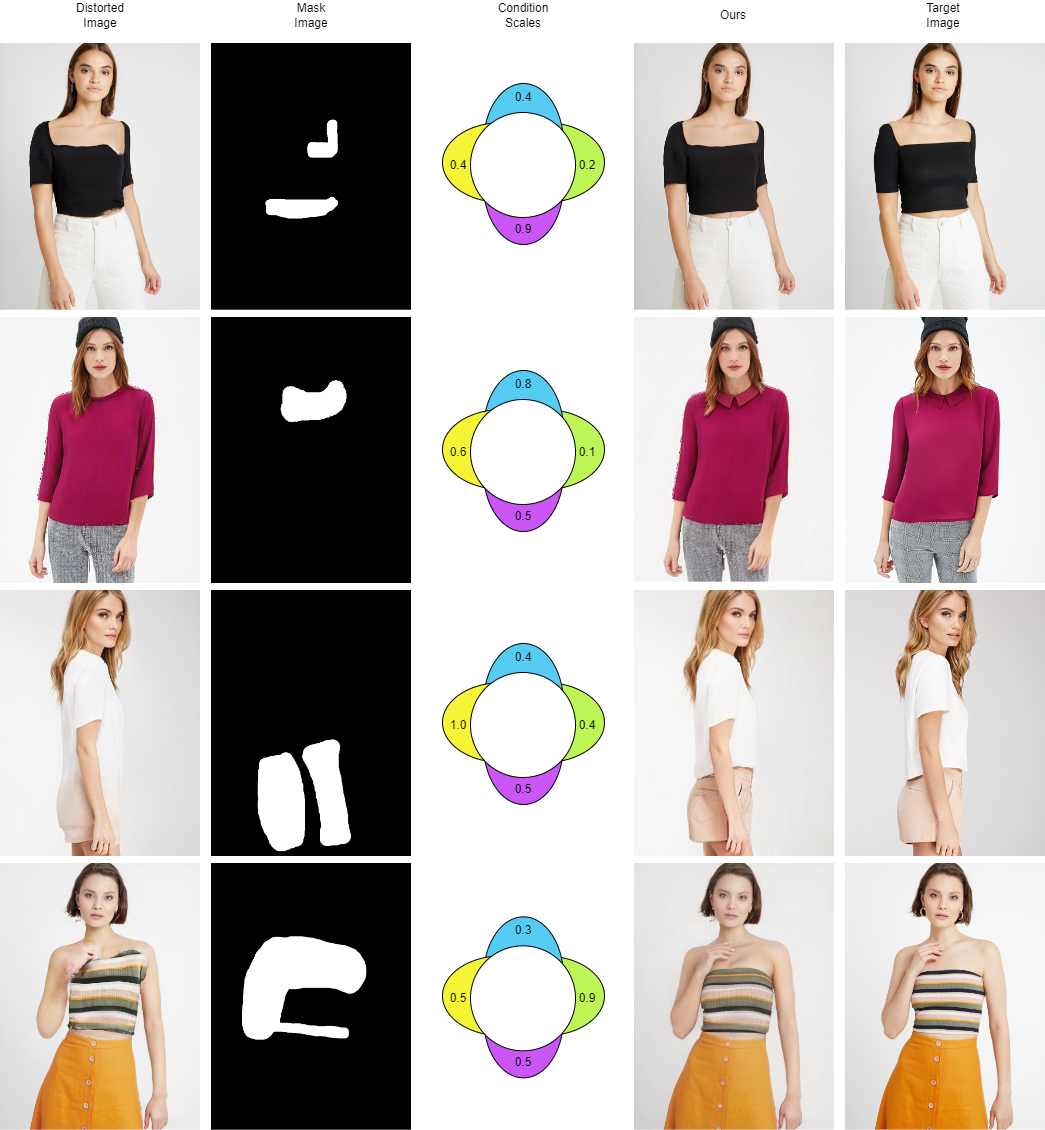}

   \caption{Artifact Removal in VTON and Pose Transfer Tasks: The 1st and 4th rows present results for the virtual try-on (VTON) task, while the 2nd and 3rd rows display results for the pose transfer task. The column for condition scales highlights the impact of different conditions. For instance, in the 2nd row, where the artifact is cloth design, the Canny condition (blue) is more effective. In the 4th row, where the artifact is human body deformation, the pose condition (green) has a better impact. In the 3rd row, with a mixture of texture and cloth design artifacts, IP-Adapter (yellow) and segmentation (purple) are more useful. }
   \label{fig11}
\end{figure}

Here we evaluate Our model with different quantitative and qualitative measurements to show the effectiveness of our proposed model. In figure \ref{fig11} two sample output of our model is shown. 
\subsection{Quantitative Evaluation}
\textbf{Metrics}. In this experiments, We evaluate the model performance with Structure Similarity Index Measure (SSIM) \cite{talebi2018learned} and Learned Perceptual Image Patch Similarity (LPIPS) \cite{zhang2018unreasonable} and Fréchet Inception Distance (FID) \cite{heusel2017gans} metrics. SSIM and LPIPS are used to calculate the reconstruction accuracy in the litreture. SSIM calculates the pixel-level image similarity, while LPIPS provides perceptual distance by employing a network trained on human judgments. FID is used to measure the realism of the generated images. It calculates the distance between the distributions of synthesized images and real images. 

\textbf{Evaluation}. To substantiate our claims,
We calculate three standard quantitative metrics for image reconstruction on the distorted outputs of HR-VTON \cite{lee2022high} and NTED \cite{ren2022neural} models, chosen as representative samples. After applying our artifact removal model to their distorted outputs, we re-calculate these metrics. As shown in table \ref{table1a}, our artifact removal model consistently improves the performance of both models, underscoring its efficacy as a post-processing enhancement for VTON and pose transfer tasks.

\begin{table}[!ht]
    \centering
    \caption{Comparison of metrics before and after applying our artifact removal model. }
    \label{table1}
    \begin{subtable}[b]{0.45\textwidth}
        \centering
        \caption{Quantitative metrics for HR-VTON \cite{lee2022high} and NTED \cite{ren2022neural}}
        \label{table1a}
        \scalebox{1}{
        \begin{tabular}{c c c c} 
         & FID↓ & SSIM↑ & LPIPS↓\\
        \hline \hline 
        HR-VTON & $49.42$& 0.65& 0.49
\\
        OURS & \textbf{36.69}& \textbf{0.84}& \textbf{0.12}
\\
        \hline
        NTED & 75.29& 0.73& 0.21
\\
        OURS & \textbf{70.40}& \textbf{0.76} & \textbf{0.17}\\
        \end{tabular}}
    \end{subtable}
    \hfill
    \begin{subtable}[b]{0.45\textwidth}
        \centering
        \caption{Qualitative metrics for VDI and DDI}
        \label{table1b}
        \scalebox{1}{
        \begin{tabular}{c c c c} 
         && Ranking↓ & Contrast↑\\
        \hline \hline 
        \multirow{3}{*}{\rotatebox{90}{VDI}}
        &Distorted & 2.92 & \textbf{5.25}\\
        &Target & \textbf{1.46} & 1.18\\
        &OURS & 1.62 & -\\
        \hline
        \multirow{3}{*}{\rotatebox{90}{DDI}}
        &Distorted & 2.82 & \textbf{7.33}\\
        &Target & \textbf{1.42} & 0.94\\
        &OURS & 1.72 & -\\
        \end{tabular}}
    \end{subtable}
\end{table}

\subsection{Qualitative Evaluation}
\textbf{Metrics}. Here we setup two experiment and report two measures; Ranking and Contrast. To compute ranking we invited 50 individuals to participate in a ranking task involving sets of three images: the original target image, a distorted version from our VDI or DDI datasets, and our model's enhanced output. Each evaluator was presented with 10 such groups and asked to rank the images based on visual quality and realism. we report the average ranking for each type of image. To measure Contrast, evaluators were shown pairs of images, one being a real or distorted image and the other our enhanced output. Each participant reviewed 10 randomly selected pairs and chose the image they deemed more realistic. We then calculated the ratio of selections favoring our model's results to those favoring the competing method.

\textbf{Evaluation}. While our quantitative evaluation demonstrates significant improvements in the technical aspects of image reconstruction,but we know these metrics sometimes fall short of capturing the full spectrum of human visual perception. This limitation is particularly relevant in fields like VTON and pose transfer, where the end results are directly interacted with by users whose satisfaction depends highly on visual realism and fidelity. To overcome the limitations of machine-based assessments, we incorporate human evaluations into our validation process based on described measures. The result of these experiments has shown in table \ref{table1b}.
The evaluation results strongly demonstrate the efficacy of our artifact removal model across both VDI and DDI datasets. The ranking metrics reveal that our model consistently outperforms the distorted versions, which indicates that participants often preferred our enhanced outputs due to their higher naturalness and realism. Moreover, the contrast metrics further underscore our model's superiority in artifact removal, with high preference ratios for our outputs over the distorted images, and nearly equal preference between our outputs and the target images. This near-random choice between our outputs and the target images highlights our model’s ability to produce results that are not only significantly better than distorted images but also comparable in quality to the original undistorted targets, underscoring its potential utility and effectiveness in real-world VTON and pose transfer applications.

\section{Implementation Details}\label{app: Exp details}
In our model's implementation, we fine-tuned the ControlNet \cite{zhang2023adding} conditions and trained the scale generator as follows:

\begin{itemize}
    \item Pose Condition Fine-tuning: Leveraged 13,678 VTON-HD images, processed over 5 epochs (10 hours) on an NVIDIA GeForce RTX 2080 Ti.
    \item Segmentation Condition Fine-tuning: Utilized 19,214 combined images from VTON-HD and DeepFashion, following the same training specifications as the pose condition.
    \item Scale Generator Training: Developed using our specially collected dataset (DDI and VDI)to adjust condition impacts based on artifact types, enhancing the model's artifact removal efficacy.
\end{itemize}

In our artifact removal process, we employed the Realistic Vision V6.0 for the base inpainting model, achieving our best results with seeds 8 and 11 for different cases. For hand and face detection in artifact identification, we utilized YOLOv5.

\subsection{Discussion}\label{app: discussion}
\begin{figure}[h]
  \centering
  \includegraphics[width=0.7\columnwidth]{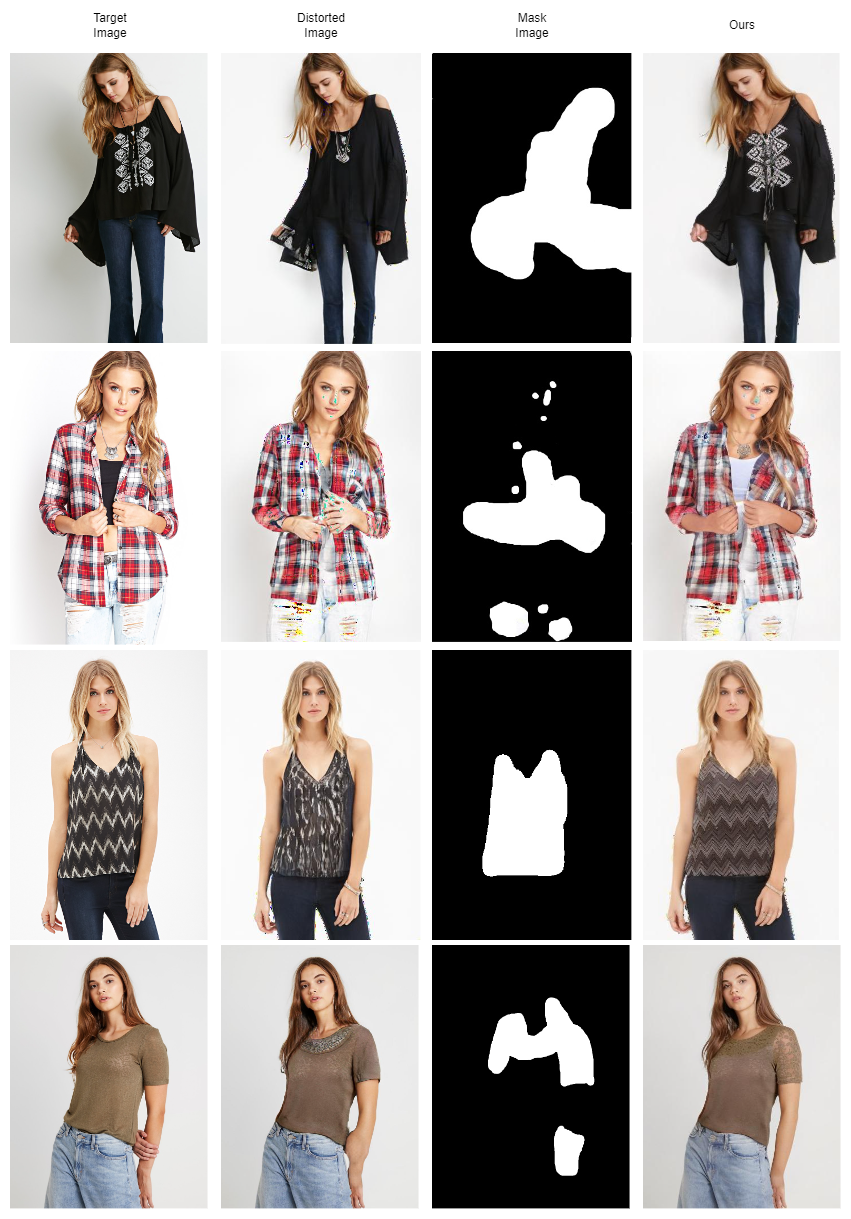}
   \caption{Limitations: Top row: Weak performance of inpainting module for cloth designs and body deformations in the Stable Diffusion model. Second row: Incorrect masking techniques leading to bad visually results. Third row: Suboptimal performance when multiple artifact types are present in an image, our model struggles to remove them correctly. Bottom row: Inadvertent inpainting of additional designs into the cloth region, highlighting a failure of the inpainting process to accurately replicate the intended content.}
   \label{fig13}
\end{figure}

As discussed in previous subsections, our method has exhibited satisfactory results in various test cases, such as text on cloth designs, striped clothing patterns, deformation in human body parts, and artifacts in the color and shape of hair and typically, when artifacts are confined to one of the classes significant improvements can be observed in distorted images. 
However, challenges arise when multiple artifacts are present. For instance, if both long hair and upper clothing worn by a human are distorted, the masked region may encompass both, posing a challenge for the inpainting module. Additionally, artifacts embedded within intricate designs of clothing can be difficult to remove, leading to suboptimal results. Also our method performance relies heavily on the performance of the inpainting module which could be an additional suboptimality. As illustrated in figure \ref{fig13}, these challenges could be the limitation of our model and future avenues with complicated training procedure and complex training data could address these problems.

\section{CONCLUSION}
\textcolor{blue}{This study introduces a novel approach to addressing the challenge of artifacts in virtual try-on (VTON) and pose transfer tasks, an area previously underexplored. We present a new artifact removal method that has shown substantial improvements, both quantitatively and qualitatively, in terms of visual quality, realism, and fidelity. Furthermore, we have curated two specialized datasets—VDI and DDI—comprising over 5000 meticulously annotated samples, aimed at advancing future research and development in this domain.}

\begin{figure}[!ht]

\end{figure}




\clearpage
\newpage
\bibliographystyle{splncs04}
\bibliography{main}

\begin{thebibliography}{10}
\providecommand{\url}[1]{\texttt{#1}}
\providecommand{\urlprefix}{URL }
\providecommand{\doi}[1]{https://doi.org/#1}

\bibitem{canny1986computational}
Canny, J.: A computational approach to edge detection. IEEE Transactions on pattern analysis and machine intelligence (6),  679--698 (1986)

\bibitem{cao2017realtime}
Cao, Z., Simon, T., Wei, S.E., Sheikh, Y.: Realtime multi-person 2d pose estimation using part affinity fields. In: Proceedings of the IEEE conference on computer vision and pattern recognition. pp. 7291--7299 (2017)

\bibitem{choi2021viton}
Choi, S., Park, S., Lee, M., Choo, J.: Viton-hd: High-resolution virtual try-on via misalignment-aware normalization. In: Proceedings of the IEEE/CVF conference on computer vision and pattern recognition. pp. 14131--14140 (2021)

\bibitem{dehghanian2023spot}
Dehghanian, Z., Saravani, S., Amirmazlaghani, M., Rahmati, M.: Spot the odd one out: Regularized complete cycle consistent anomaly detector gan. arXiv preprint arXiv:2304.07769  (2023)

\bibitem{fele2022c}
Fele, B., Lampe, A., Peer, P., Struc, V.: C-vton: Context-driven image-based virtual try-on network. In: Proceedings of the IEEE/CVF winter conference on applications of computer vision. pp. 3144--3153 (2022)

\bibitem{ge2021parser}
Ge, Y., Song, Y., Zhang, R., Ge, C., Liu, W., Luo, P.: Parser-free virtual try-on via distilling appearance flows. In: Proceedings of the IEEE/CVF conference on computer vision and pattern recognition. pp. 8485--8493 (2021)

\bibitem{gonzalez2009digital}
Gonzalez, R.C., Woods, R.E.: Digital Image Processing. Prentice Hall (2009)

\bibitem{goodfellow2014generative}
Goodfellow, I., Pouget-Abadie, J., Mirza, M., Xu, B., Warde-Farley, D., Ozair, S., Courville, A., Bengio, Y.: Generative adversarial nets. Advances in neural information processing systems  \textbf{27} (2014)

\bibitem{gou2023taming}
Gou, J., Sun, S., Zhang, J., Si, J., Qian, C., Zhang, L.: Taming the power of diffusion models for high-quality virtual try-on with appearance flow. In: Proceedings of the 31st ACM International Conference on Multimedia. pp. 7599--7607 (2023)

\bibitem{han2018viton}
Han, X., Wu, Z., Wu, Z., Yu, R., Davis, L.S.: Viton: An image-based virtual try-on network. In: Proceedings of the IEEE conference on computer vision and pattern recognition. pp. 7543--7552 (2018)

\bibitem{he2016deep}
He, K., Zhang, X., Ren, S., Sun, J.: Deep residual learning for image recognition. In: Proceedings of the IEEE conference on computer vision and pattern recognition. pp. 770--778 (2016)

\bibitem{heusel2017gans}
Heusel, M., Ramsauer, H., Unterthiner, T., Nessler, B., Hochreiter, S.: Gans trained by a two time-scale update rule converge to a local nash equilibrium. Advances in neural information processing systems  \textbf{30} (2017)

\bibitem{ho2020denoising}
Ho, J., Jain, A., Abbeel, P.: Denoising diffusion probabilistic models. Advances in neural information processing systems  \textbf{33},  6840--6851 (2020)

\bibitem{kang2023scaling}
Kang, M., Zhu, J.Y., Zhang, R., Park, J., Shechtman, E., Paris, S., Park, T.: Scaling up gans for text-to-image synthesis. In: Proceedings of the IEEE Conference on Computer Vision and Pattern Recognition (CVPR). pp. 10124--10134 (2023)

\bibitem{lee2022high}
Lee, S., Gu, G., Park, S., Choi, S., Choo, J.: High-resolution virtual try-on with misalignment and occlusion-handled conditions. In: European Conference on Computer Vision. pp. 204--219. Springer (2022)

\bibitem{li2022blip}
Li, J., Li, D., Xiong, C., Hoi, S.: Blip: Bootstrapping language-image pre-training for unified vision-language understanding and generation. In: International Conference on Machine Learning. pp. 12888--12900. PMLR (2022)

\bibitem{li2019selfcorrection}
Li, P., Xu, Y., Wei, Y., Yang, Y.: Self-correction for human parsing. arXiv preprint arXiv:1910.09777  (2019)

\bibitem{li2023virtual}
Li, Z., Wei, P., Yin, X., Ma, Z., Kot, A.C.: Virtual try-on with pose-garment keypoints guided inpainting. In: Proceedings of the IEEE/CVF International Conference on Computer Vision. pp. 22788--22797 (2023)

\bibitem{liuLQWTcvpr16DeepFashion}
Liu, Z., Luo, P., Qiu, S., Wang, X., Tang, X.: Deepfashion: Powering robust clothes recognition and retrieval with rich annotations. In: Proceedings of IEEE Conference on Computer Vision and Pattern Recognition (CVPR) (June 2016)

\bibitem{ma2017pose}
Ma, L., Jia, X., Sun, Q., Schiele, B., Tuytelaars, T., Van~Gool, L.: Pose guided person image generation. Advances in neural information processing systems  \textbf{30} (2017)

\bibitem{morelli2023ladi}
Morelli, D., Baldrati, A., Cartella, G., Cornia, M., Bertini, M., Cucchiara, R.: Ladi-vton: Latent diffusion textual-inversion enhanced virtual try-on. arXiv preprint arXiv:2305.13501  (2023)

\bibitem{morelli2022dress}
Morelli, D., Fincato, M., Cornia, M., Landi, F., Cesari, F., Cucchiara, R.: Dress code: High-resolution multi-category virtual try-on. In: Proceedings of the IEEE/CVF Conference on Computer Vision and Pattern Recognition. pp. 2231--2235 (2022)

\bibitem{redmon2016you}
Redmon, J., Divvala, S., Girshick, R., Farhadi, A.: You only look once: Unified, real-time object detection. In: Proceedings of the IEEE conference on computer vision and pattern recognition. pp. 779--788 (2016)

\bibitem{ren2022neural}
Ren, Y., Fan, X., Li, G., Liu, S., Li, T.H.: Neural texture extraction and distribution for controllable person image synthesis. In: Proceedings of the IEEE/CVF Conference on Computer Vision and Pattern Recognition. pp. 13535--13544 (2022)

\bibitem{ren2021combining}
Ren, Y., Wu, Y., Li, T.H., Liu, S., Li, G.: Combining attention with flow for person image synthesis. In: Proceedings of the 29th ACM International Conference on Multimedia. pp. 3737--3745 (2021)

\bibitem{ren2020deep}
Ren, Y., Yu, X., Chen, J., Li, T.H., Li, G.: Deep image spatial transformation for person image generation. In: Proceedings of the IEEE/CVF Conference on Computer Vision and Pattern Recognition. pp. 7690--7699 (2020)

\bibitem{rombach2022high}
Rombach, R., Blattmann, A., Lorenz, D., Esser, P., Ommer, B.: High-resolution image synthesis with latent diffusion models. In: Proceedings of the IEEE/CVF conference on computer vision and pattern recognition. pp. 10684--10695 (2022)

\bibitem{sarkar2021style}
Sarkar, K., Golyanik, V., Liu, L., Theobalt, C.: Style and pose control for image synthesis of humans from a single monocular view. arXiv preprint arXiv:2102.11263  (2021)

\bibitem{tabatabaei2024no}
Tabatabaei, A., Dehghanian, Z., Movaghatian, N., Amirmazlaghani, M.: No more blah-blah: Embracing real text in the image synthesis world. In: The Second Tiny Papers Track at ICLR (2024)

\bibitem{talebi2018learned}
Talebi, H., Milanfar, P.: Learned perceptual image enhancement. In: 2018 IEEE international conference on computational photography (ICCP). pp. 1--13. IEEE (2018)

\bibitem{wang2018toward}
Wang, B., Zheng, H., Liang, X., Chen, Y., Lin, L., Yang, M.: Toward characteristic-preserving image-based virtual try-on network. In: Proceedings of the European conference on computer vision (ECCV). pp. 589--604 (2018)

\bibitem{ye2023ip-adapter}
Ye, H., Zhang, J., Liu, S., Han, X., Yang, W.: Ip-adapter: Text compatible image prompt adapter for text-to-image diffusion models  (2023)

\bibitem{zhang2021pise}
Zhang, J., Li, K., Lai, Y.K., Yang, J.: Pise: Person image synthesis and editing with decoupled gan. In: Proceedings of the IEEE/CVF Conference on Computer Vision and Pattern Recognition. pp. 7982--7990 (2021)

\bibitem{zhang2020human}
Zhang, J., Liu, X., Li, K.: Human pose transfer by adaptive hierarchical deformation. In: Computer Graphics Forum. vol.~39, pp. 325--337. Wiley Online Library (2020)

\bibitem{zhang2023adding}
Zhang, L., Rao, A., Agrawala, M.: Adding conditional control to text-to-image diffusion models. In: Proceedings of the IEEE/CVF International Conference on Computer Vision. pp. 3836--3847 (2023)

\bibitem{zhang2018unreasonable}
Zhang, R., Isola, P., Efros, A.A., Shechtman, E., Wang, O.: The unreasonable effectiveness of deep features as a perceptual metric. In: Proceedings of the IEEE conference on computer vision and pattern recognition. pp. 586--595 (2018)

\bibitem{zhu2023tryondiffusion}
Zhu, L., Yang, D., Zhu, T., Reda, F., Chan, W., Saharia, C., Norouzi, M., Kemelmacher-Shlizerman, I.: Tryondiffusion: A tale of two unets. In: Proceedings of the IEEE/CVF Conference on Computer Vision and Pattern Recognition. pp. 4606--4615 (2023)

\end{thebibliography}

\newpage

\appendix
\label{app: further results}

\clearpage


\end{document}